\documentclass[10pt,letterpaper]{article}

\usepackage{cogsci}

\cogscifinalcopy % Uncomment this line for the final submission 

\usepackage{pslatex}
\usepackage{apacite}
\usepackage{float} % Roger Levy added this and changed figure/table
\usepackage{graphicx}
\usepackage{tabularx}
\usepackage{balance}

\usepackage{lmodern}
\usepackage{times}

\usepackage{algorithm}
\usepackage{algorithmic}
\usepackage{amsfonts}
\usepackage{tikz}
\usepackage{xcolor}
\usepackage{amsmath}
\usepackage{amsthm}
 
\DeclareMathOperator*{\E}{\mathbb{E}}

\newcommand{\Fb}{\F_{\beta}}
\newcommand{\qb}{q_{\beta}}

\newcommand{\bl}{\beta_l}
\newcommand{\el}{\varepsilon_l}
\newcommand{\mhat}{\hat{m}}
\newcommand{\Mhat}{\hat{M}}

\newcommand{\mw}{\mhat_w}

 \newcommand{\mbf}[1]{\mathbf{#1}}

\newcommand{\w}{\mbf{w}}

\newcommand{\cU}{\mathcal{U}}
\newcommand{\F}{\mathcal{F}}

\newcommand{\eqn}[1]{\begin{equation} #1 \end{equation}}

\newcommand{\Sim}{\mathrm{sim}}

\renewcommand{\eqref}[1]{equation~(\ref{#1})}

\newcommand{\figref}[1]{Figure~\ref{#1}}

\usepackage{fancyhdr}
\title{Semantic categories of artifacts and animals reflect efficient coding}

\author{
{\large \bf Noga Zaslavsky$^{1,2}$ (noga.zaslavsky@mail.huji.ac.il)}\\
{\large \bf Terry Regier$^{2,3}$ (terry.regier@berkeley.edu)}\\
{\large \bf Naftali Tishby$^{1,4}$ (tishby@cs.huji.ac.il)}\\
{\large \bf Charles Kemp$^5$ (c.kemp@unimelb.edu.au)}
\AND
$^{1}$Edmond and Lily Safra Center for Brain Sciences,  Hebrew University, Jerusalem 9190401, Israel\\
$^{2}$Department of Linguistics, University of California, Berkeley, CA 94720 USA\\
$^{3}$Cognitive Science Program, University of California, Berkeley, CA 94720 USA\\
$^{4}$Benin School of Computer Science and Engineering,  Hebrew University, Jerusalem 9190401, Israel\\
$^{5}$School of Psychological Sciences, University of Melbourne, Parkville, Victoria 3010, Australia
}

\begin{document}

\maketitle

\begin{abstract}
It has been argued that semantic categories across languages reflect pressure for efficient communication. Recently, this idea has been cast in terms of a general information-theoretic principle of efficiency, the Information Bottleneck (IB) principle, and it has been shown that this principle accounts for the emergence and evolution of named color categories across languages, including soft structure and patterns of inconsistent naming.  However, it is not yet clear to what extent this account generalizes to semantic domains other than color. Here we show that it generalizes to two qualitatively different semantic domains: names for containers, and for animals. First, we show that container naming in Dutch and French is near-optimal in the IB sense, and that IB broadly accounts for soft categories and inconsistent naming patterns in both languages. Second, we show that a hierarchy of animal categories derived from IB captures cross-linguistic tendencies in the growth of animal taxonomies. Taken together, these findings suggest that fundamental information-theoretic principles of efficient coding may shape semantic categories across languages and across domains.

\textbf{Keywords:} 
information theory; language evolution; semantic typology; categories
\end{abstract}

\thispagestyle{fancy}

\section{Introduction}

Cross-linguistic studies in several semantic domains, such as kinship, color, and numeral systems, suggest that word meanings are adapted for efficient communication (see~\citeNP{Kemp2018} for a review). However, until recently it had remained largely unknown to what extent this proposal can account for soft semantic categories and inconsistent naming, that could appear to pose a challenge to the notion of efficiency, and how pressure for efficiency may relate to language evolution.
Recently \citeauthor{Zaslavsky2018} (2018; henceforth ZKRT) addressed these open questions by grounding the notion of efficiency in a general information-theoretic principle, the Information Bottleneck (IB; \citeNP{Tishby1999}). ZKRT tested this formal approach in the domain of color naming and showed that the IB principle: (1) accounts to a large extent for cross-language variation in color naming; (2) provides a theoretical explanation for why observed patterns of inconsistent naming and soft semantic categories may be efficient; and (3) suggests a possible evolutionary process that roughly recapitulates Berlin and Kay's (1969) discrete implicational hierarchy \nocite{Berlin1969} while also accounting for continuous aspects of color category evolution. However, it is not yet clear to what extent these results may generalize to other semantic domains, especially those that are fundamentally unlike color.

Here we test the generality of this theoretical account by considering two additional semantic domains: artifacts and animals. These domains are of particular interest in this context because they are qualitatively different from color, they have not previously been comprehensively addressed in terms of efficient communication, and at the same time it is possible to apply to them the same communication model that has previously been used to account for color naming.

First, we consider naming patterns for household containers.
This is a semantic domain in which categories are known to overlap and generate inconsistent naming patterns~\cite{Ameel2005,Ameel2009}.
Although it has previously been shown that container naming in English, Spanish, and Chinese is efficient compared to a large set of hypothetical naming systems~\cite{Xu2016}, that demonstration did not consider the full probability distribution of names produced by different speakers, did not explicitly contrast monolingual and bilingual speakers, and was based on a smaller set of stimuli than we consider here.
In this work we show that the full container-naming distribution in Dutch and French, including overlapping and inconsistent naming patterns, across a large set of stimuli, both in monolinguals and bilinguals, is near-optimally efficient in the IB sense.

Second, we test the evolutionary account of ZKRT
in the case of animal categories.
By analogy with Berlin and Kay's implicational hierarchy of color terms, \citeA{Brown1984} proposed an implicational hierarchy for the evolution of animal taxonomies based on cross-language comparison.
We show that aspects of this hierarchy are captured by a sequence of efficient animal-naming systems along the IB theoretical limit. Our results also support the view that both perceptual and functional features shape animal categories across languages~\cite{Malt1995,Kemp2018}. 

The remainder of this paper proceeds as follows. First, we review the theoretical framework and formal predictions on which we build. We then present two studies that apply this approach to the aforementioned semantic domains.

\section{Theoretical framework and predictions}

\begin{figure}[t]
\centering
\includegraphics[width=0.85\linewidth]{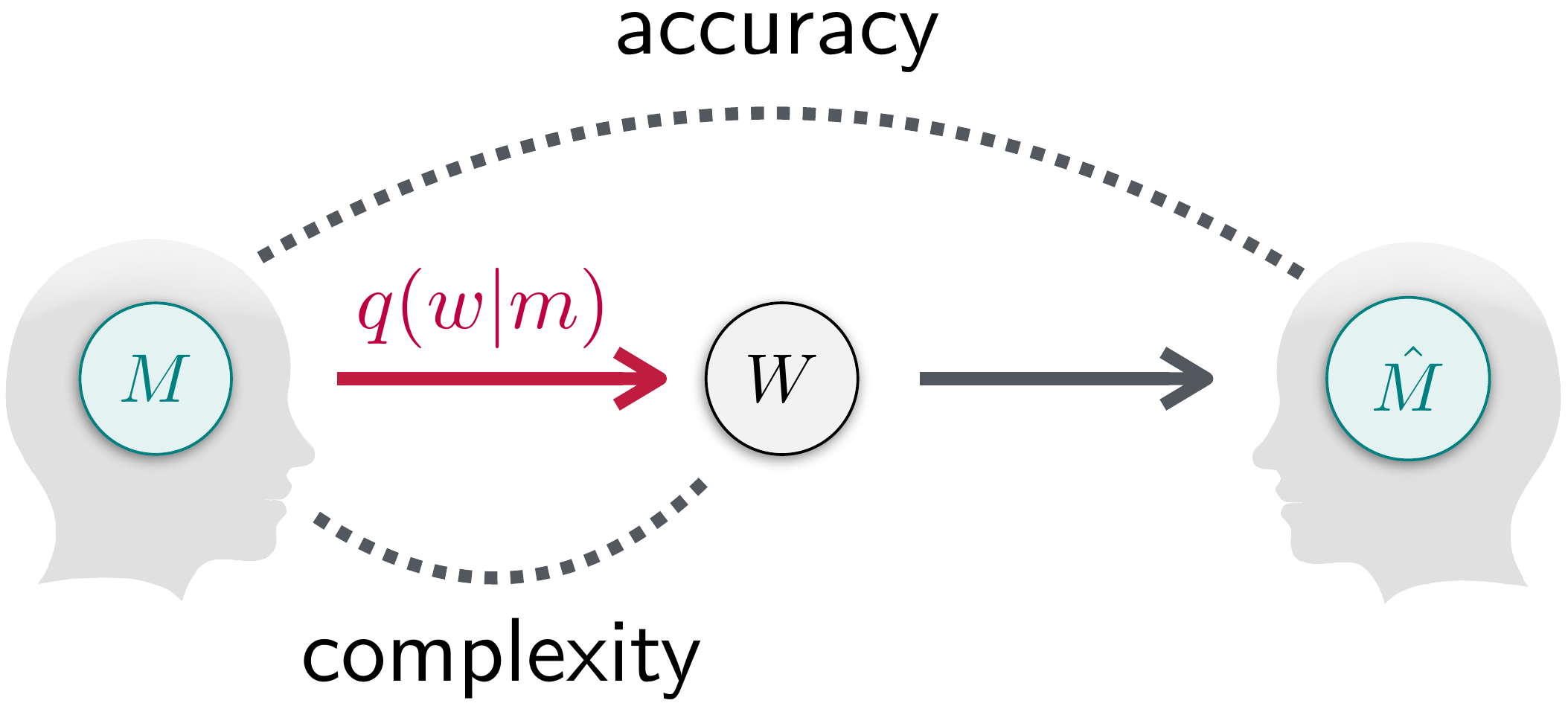}
\caption{Communication model adapted from ZKRT.
A speaker communicates a meaning $M$ by encoding it into a word $W$ according to a naming distribution $q(w|m)$. This word is then interpreted by the listener as $\Mhat$. Complexity is a property of the mapping from meanings to words, and accuracy is determined by the similarity between $M$ and $\Mhat$.}
\label{fig:comm-model}
\end{figure}

We consider here the theoretical framework proposed by ZKRT,
which is based on a simplified interaction between a speaker and a listener (\figref{fig:comm-model}), formulated in terms of Shannon's~\citeyear{Shannon1948} communication model. The speaker communicates a meaning $m$, sampled from $p(m)$, by encoding it into a word $w$, generated from a naming (or encoder) distribution $q(w|m)$. The listener then tries to reconstruct from $w$ the speaker's intended meaning. We denote the reconstruction by $\mw$, and assume it is obtained by a Bayesian listener.\footnote{The reconstruction of a Bayesian listener with respect to a given naming distribution is defined by $\mw = \sum_m q(m|w)m$.} These meanings, $m$ and $\mw$, are taken to be mental representations of the environment, defined by distributions over a set $\cU$ of relevant features. For example, if communication is about colors, then $\cU$ may be grounded in a perceptual color space, and each color would be mentally represented as a distribution over this space.

Under these assumptions, efficient communication systems are those naming distributions that optimize the Information Bottleneck (IB; \citeNP{Tishby1999}) tradeoff between the complexity and accuracy of the lexicon.
Formally, complexity is measured by the mutual information between meanings and words, i.e.:
\eqn{
I_q(M;W) = \sum_{m,w} p(m) q(w|m) \log \frac{q(w|m)}{q(w)}\,,\label{eq:Ix}
}
which roughly corresponds to the number of bits used to encode meanings into words. Accuracy is inversely related to the discrepancy between $m$ and $\mw$, measured by the expected Kullback--Leibler (KL) divergence between them:
\eqn{
\mathbb{E}_{q}[D[m\|\mw]] = \E_{\substack{m\sim p(m)\\w\sim q(w|m)}}\left[\sum_{u\in\cU} m(u)\log \frac{m(u)}{\mw(u)}\right]\,.\label{eq:D}
}
Accuracy is defined
by  $I_q(W;U)=\mathbb{E}_{q}[D[\mw\|m_0]]$, where $m_0$ is the prior representation before knowing $w$, and maximizing accuracy amounts to minimizing \eqref{eq:D}.\footnote{See~\cite{Zaslavsky2018} 
for detailed explanation.}

\begin{figure}[t!]
\centering
\includegraphics[width=.95\linewidth,trim={0 1em 0 1em},clip]{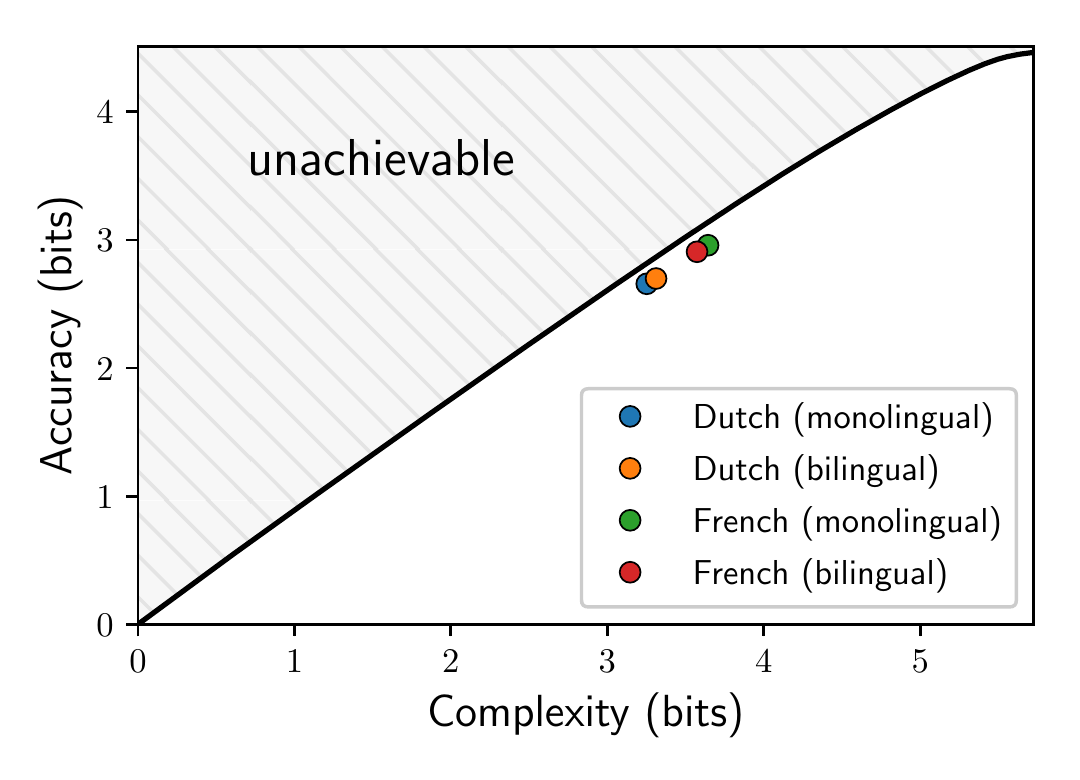}
\caption{The black curve is the IB theoretical limit of efficiency for container naming, obtained by varying $\beta$. Points above this curve cannot be achieved. Complexity and accuracy tradeoffs in the four naming conditions are near-optimal.}
\label{fig:containers_info_plane}
\end{figure}

Achieving maximal accuracy may require a highly complex system, while minimizing complexity will result in a non-informative system. Efficient systems are thus pressured to balance these two competing goals by minimizing the IB objective function,
\eqn{
\Fb[q] = I_q(M;W) - \beta I_q(W;U)\,,\label{eq:IBP}
}
where $\beta\ge0$ controls the efficiency tradeoff.
The optimal systems, $\qb(w|m)$, achieve the minimal value of \eqref{eq:IBP} given $\beta$, denoted by $\Fb^*$, and evolve as $\beta$ gradually shifts from $0$ to $\infty$. Along this trajectory they become more fine-grained and complex, while attaining the maximal achievable accuracy for their level of complexity. This set of optimal systems defines the theoretical limit of efficiency (see \figref{fig:containers_info_plane}).

If languages are pressured to be efficient in the IB sense, then for a given language $l$ with naming system $q_l(w|m)$, two predictions are made.
(1) Deviation from optimality, or \textit{inefficiency}, should be small. This is measured by $\el = \tfrac{1}{\bl}(\F_{\bl}[q_l] - \F_{\bl}^*)$, where $\bl$ is estimated such that $\el$ is minimized.
(2) The \textit{dissimilarity} between $q_l$ and the corresponding IB system, $q_{\bl}$, should be small. This is evaluated by a dissimilarity measure (gNID) proposed by ZKRT.
In addition, ZKRT suggested that languages evolve along a trajectory that is pressured to remain near the theoretical limit.

\begin{figure*}[t!]
\centering
\includegraphics[width=\textwidth]{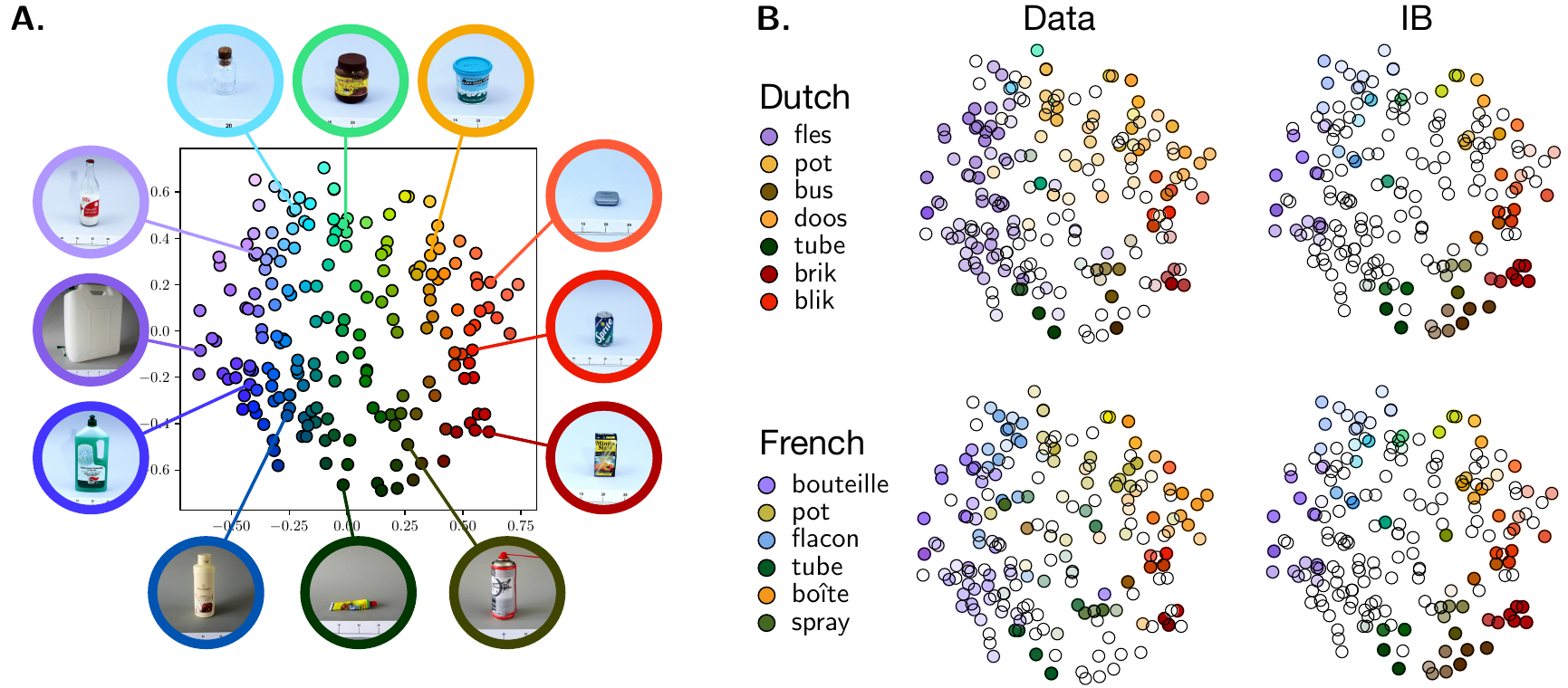}
\caption{
\textbf{A}. Two dimensional nMDS embedding and color coding of the containers stimulus set used by \citeA{White2017}. Images show a few examples.
\textbf{B}. Monolingual naming distributions for Dutch (upper left) and French (lower left), together with their corresponding IB systems (right column), are visualized over the 2D embedding shown in (A). Each color corresponds to the color centroid of a container category, $w$, based on the color map in (A). Colors show category probabilities above 0.4, and color intensities reflect the values between 0.4 and 1. White dots correspond to containers for which no category is used with probability above 0.4. Legend for each language shows only major terms.
}
\label{fig:containers_nmds}
\end{figure*}

These predictions were previously supported by evidence from the domain of color naming.
To apply this approach to other domains, i.e.\ to instantiate the general communication model, two components must be specified: a \textit{meaning space}, which is the set of meanings the speaker may communicate; and a prior, $p(m)$, also referred to as a \textit{need distribution}~\cite{Regier2015}, since it determines the frequency with which each meaning needs to be communicated.
In the following sections we present two studies that follow this approach and test its predictions in qualitatively different semantic domains.

\section{Study I: Container names}

The goal of this experiment is to test the theoretical predictions derived from IB in the case of container naming. It is not clear whether previous findings for color would generalize to this case for several reasons. First, the representation of artifacts is likely to involve more than just a few basic perceptual features, unlike color. Second, categories in this domain are believed to be strongly shaped by adaptation to changes in the environment~\cite{Malt1999}.
At the same time, container categories tend to overlap, as in the case of color categories, posing a similar theoretical challenge to explain this observation in terms of communicative efficiency. Finally, the bilingual lexicon in this domain has been extensively studied, and it has been shown that bilingual naming patterns tend to converge~\cite{Ameel2005,Ameel2009}.
However, it is not yet clear whether this convergence, or compromise, comes at a cost in communicative efficiency, or whether it may actually be formalized and explained in terms of efficiency.

\subsubsection{Data.}

To address these open questions, we consider sorting and naming data collected by \citeA{White2017}, relative to a stimulus set of 192 images of household containers (see \figref{fig:containers_nmds}A for examples). This set is substantially larger than those used in previous container-naming studies (e.g.~\citeNP{Malt1999,Ameel2005}), thus providing a better representation of this semantic domain.
In the naming task, 32 Dutch and 30 French monolingual speakers, as well as 30 bilingual speakers, were asked to provide names for the containers in the stimulus set. Bilingual participants performed the task once in each language. The container-naming distribution in each of the four conditions (language $\times$ linguistic status) is defined by the proportion of participants in that condition that used the word $w$ to describe a container $c$.
A separate sorting task was performed by 65 Dutch speakers, who were asked to organize all containers into piles based on their overall qualities.
Participants were also allowed to form higher-level clusters by grouping piles together.
\citeA{White2017} evaluated the similarity between two containers, denoted here by $\Sim(c,c')$, based on the number of participants that placed them in the same pile or cluster (see~\citeNP{White2017} for detail).
In both tasks, participants were instructed not to take into account the content of the object (e.g., water).

\subsubsection{Model.}

We ground the meaning space in the similarity data, following a related approach proposed by~\citeA{Regier2015} and~\citeA{Xu2016}. While these data are from Dutch speakers, there are only minor differences in perceived similarities among speakers of different languages~\cite{Ameel2005}. Therefore, we assume that these similarity judgments reflect a shared underlying perceptual representation of this domain. We take $\cU$ to be the set of containers in the stimulus set, and define the mental representation of each container $c$ by the similarity-based distribution it induces over the domain,
$m_c(u) \propto \exp\left(\gamma\cdot \Sim(c,u)\right)$,
where $\gamma^{-1}$ is taken to be the empirical standard deviation of $\Sim(c,u)$.
In contrast with the case of color, in which these mental representations were grounded in a standard perceptual space, here there is no standard perceptual space for containers, and so our assumed underlying perceptual representation requires further validation, which we leave for future work.
We define the need distribution, $p(m_c)$, by averaging together the least informative (LI) priors for the different languages, as proposed by ZKRT.
We used only the monolingual data for this purpose, and regularized the resulting prior by adding  $\epsilon=0.001$ to it and renormalizing. 

\subsection{Results}

We estimated the theoretical limit of efficiency for container naming by applying the IB method~\cite{Tishby1999}, as ZKRT did in the case of color naming, here with $1500$ values of $\beta\in[0,1024]$. We evaluated the empirical complexity and accuracy in the four naming conditions by entering the corresponding naming distributions in the equations for $I_q(M;W)$ and $I_q(W;U)$. The results are shown in~\figref{fig:containers_info_plane} and Table~\ref{tab:quantitative}. It can be seen that container naming in Dutch and French lie near theoretical limit, both for monolinguals and bilinguals, and that bilinguals achieve similar levels of efficiency as monolinguals (Table~\ref{tab:quantitative}). In all four cases, the corresponding IB solution is at $\beta_l\approx1.2$, suggesting that there is only a weak preference for accuracy over complexity in this domain, as also found for color naming.

Consistent with the empirical observations of convergence in the bilingual lexicon, the complexity-accuracy tradeoffs in bilinguals are closer to each other (\figref{fig:containers_info_plane}, orange and red dots) compared to the monolingual tradeoffs (\figref{fig:containers_info_plane}, blue and green dots). This may be explained by a need to reduce the complexity of maintaining two naming systems simultaneously, while achieving monolingual-like levels of efficiency in each language. To test this possibility, we compared two joint French-Dutch systems that bilinguals may employ: one that randomly selects one of the two monolingual systems to name objects, and another that randomly selects one of the two bilingual systems. We found a $0.16\%$ reduction in the complexity of the joint bilingual system compared to the joint monolingual system. Although this is a small effect, it may accumulate across domains to have a substantial impact.
In addition, our simple calculation did not take into account similar word forms, which may also reduce complexity~\cite{Ameel2005}. Thus, this finding suggests that the convergence in the bilingual lexicon may be shaped, at least in part, by pressure for efficiency.

The remainder of our analysis focuses on the monolingual systems, as they are more distinct and presumably more representative of each language.
To get a precise sense of how challenging it may be to reach the observed levels of efficiency, we compared the actual naming systems to a set of hypothetical systems that preserve some of their statistical structure. This set was constructed by fixing the conditional distributions of words, while shifting how they are used by applying a random permutation of the containers. For each language we constructed $10,000$ such hypothetical systems. Table~\ref{tab:quantitative} shows that these hypothetical systems are substantially less efficient than the actual systems, and are also less similar to the IB systems. In fact, both languages achieve better (lower) scores than all of their hypothetical variants, providing a precise sense in which they are near-optimal according to IB. One possible concern is that this outcome may be a result of the LI prior, which was fitted to the naming data. To address this, we repeated this analysis with a uniform need distribution. The results in that case are similar (not shown), although as expected the fit to the actual systems is not as good compared to the LI prior.

\begin{figure*}[t!]
\centering
\includegraphics[width=\textwidth]{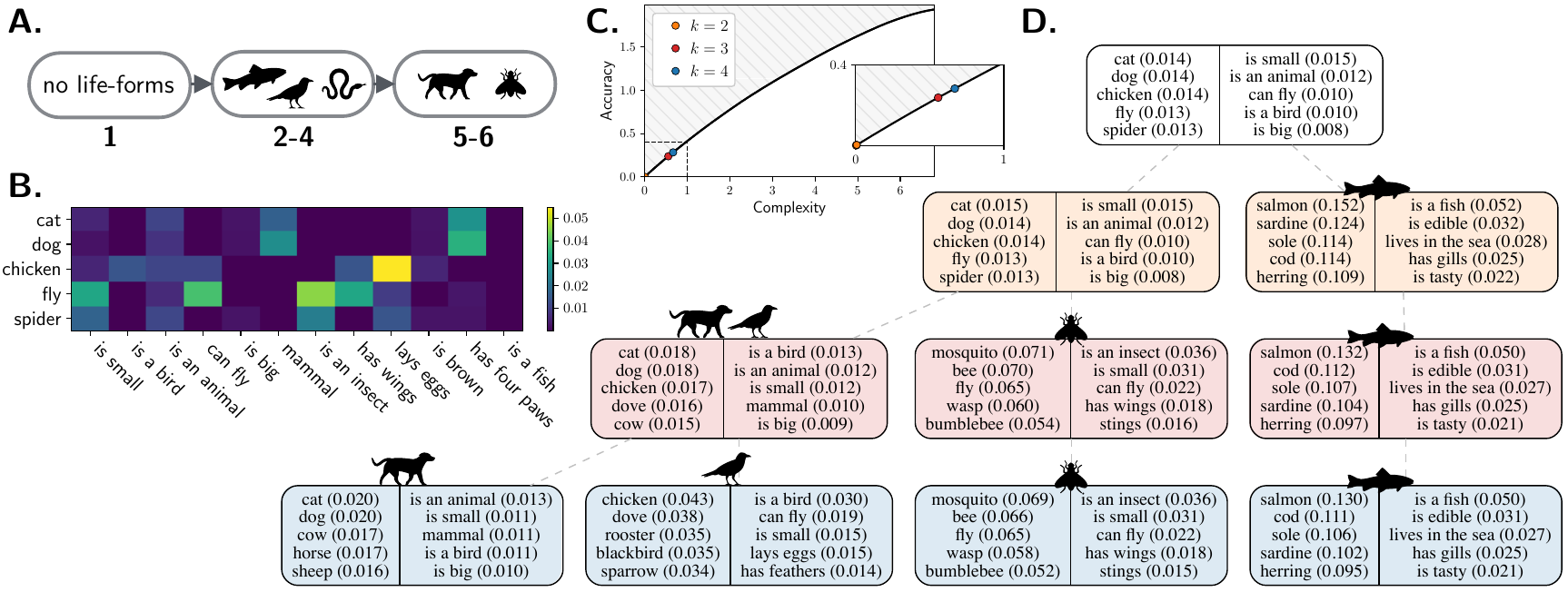}
\caption{
\textbf{A}. Brown's~\citeyear{Brown1984} proposed hierarchy for animal categories.
\textbf{B}. Subset of the conditional probabilities of features (columns) given animal classes (rows), for the 5 most familiar classes and 12 most frequently generated features.
\textbf{C}. Theoretical limit for animal naming. Colored dots along the curve correspond to the systems shown in (D), with $k=2,3,4$ categories.
\textbf{D}.~Animal category hierarchy derived from IB. Each level corresponds to an IB system. Each box corresponds to a category, which is represented by its top five classes (left) and features (right) and their probabilities given the category.
}
\label{fig:animals}
\end{figure*}

The low dissimilarity scores for the actual languages, shown in Table~\ref{tab:quantitative}, suggest that the observed soft category structure in this domain may also be accounted for by the IB systems. This is indeed supported by a fine-grained comparison between the naming distribution in both languages and their corresponding IB systems. To see this, we embedded the 192 containers in a 2-dimensional space by applying non-metric multidimensional scaling (nMDS) with respect to the similarity data, similar to~\citeA{Ameel2009}. This was done using the scikit-learn package in Python. We initialized the nMDS procedure with a solution for the standard metric MDS that achieved the best fit to the similarity data out of 50 solutions generated with random initial conditions. For visualization purposes, we assigned a unique color to each container. The resulting 2D embedding and color coding of the containers stimulus set are shown in \figref{fig:containers_nmds}A.

\begin{table}[t!]
\centering
\vspace{-.5em}
\caption{Evaluation of the IB container-naming model. Lower values indicate a better fit of the model. Values for hypothetical systems are averages $\pm$SD over $10,000$ systems.}
\label{tab:quantitative}
\vskip 0.12in
\begin{tabular}{ll|llc}
&                & Inefficiency     & Dissimilarity\\
\hline
Dutch
& monolingual    & 0.16 		    & 0.11              \\
& bilingual      & 0.17 		    & 0.12              \\
& hypothetical   & 0.29	($\pm$0.02)	& 0.59 ($\pm$0.05)  \\

\hline
French
& monolingual	& 0.18 	        	& 0.11              \\
& bilingual	    & 0.17 	        	& 0.09              \\
& hypothetical	& 0.31 ($\pm$0.01)  & 0.56 ($\pm$0.06)  \\
\hline
\end{tabular} 
\end{table}

The monolingual systems in Dutch and French are shown in~\figref{fig:containers_nmds}B, together with their corresponding IB systems. These two IB systems are very similar, although not identical, which is not surprising given that the naming patterns in Dutch and French are fairly similar.
Both the actual systems and the IB systems exhibit soft category structure and similar patters of inconsistent naming, as shown by the white dots. In addition, since each category is colored according to its centroid, similarity between the category colors together with their spatial distribution reflect the similarity between the full naming distributions. 
For example, the IB systems have a category that is similar to \textit{fles} and \textit{bouteille}, as well as a category that is similar to \textit{doos} and \textit{bo\^ite} in Dutch and French respectively, although these categories in the IB systems are a bit narrower. The IB systems also capture the category \textit{tube} quite well in both languages. However, there are also some apparent discrepancies. For example, the distinction between \textit{bouteille} and \textit{flacon} in French is reflected in both IB systems, although Dutch does not have the same pattern in this case~\cite{Ameel2005}.

This analysis shows that efficiency constraints may to a substantial extent explain the container-naming distribution in Dutch and French, including soft category boundaries and inconsistent naming observed empirically, both in monolinguals and bilinguals. It thus supports the hypothesis that a drive for information-theoretic efficiency shapes word meanings across languages and across semantic domains. However, since this analysis is based only on two closely related languages, we were not able to test how well the results for this domain generalize across languages.
Important directions for future research include testing whether these results generalize to other, preferably unrelated, languages, and further testing the extent to which the convergence in the bilingual lexicon is influenced by pressure for efficiency.
The next section focuses on another semantic domain for which we are able to obtain broader cross-linguistic evidence.

\section{Study II: Folk biology}

Cross-language variation and universal patterns in animal taxonomies have been extensively documented and studied~\cite{Berlin1992}, however this domain has not yet been approached in terms of efficient communication. 
By analogy with Berlin and Kay's theory, \citeA{Brown1984} proposed an implicational hierarchy for animal terms, based on data from 144 languages.
Brown identified six stages for animal taxonomies, as illustrated in~\figref{fig:animals}A. Languages at the first stage do not have any lexical representation for life-forms. Languages at stages 2-4 add terms for \textit{fish}, \textit{bird} and \textit{snake}, but Brown does not argue for any particular order for these categories. Terms for \textit{mammal} and \textit{wug} (``worm-bug'', referring in addition to small insects) are added in stages 5 and 6, again with no implied order.
Much of the data analyzed in this domain is not fine-grained, and Brown's proposal has been criticized~\cite{Randall1984} mainly due to lack of sufficiently accurate data. Nonetheless, his observations can be considered as a rough approximation of cross-linguistic tendencies in this semantic domain. Therefore, in this work we aim at testing whether broad cross-linguistic patterns, as summarized by Brown's proposal, can be accounted for in terms of pressure for efficiency. More specifically, our goal is to derive from the IB principle a trajectory of efficient animal-naming systems, analogous to ZKRT's trajectory for color, and to compare this trajectory to the naming patterns reported by Brown. However, unlike previous comparisons to IB optima, due to the nature of available data, here we only attempt to make coarse comparisons.

To derive a trajectory of efficient animal-naming systems, we first need to specify the communication model in this domain.
We ground the representations of animals in high-level, human-generated features.
Specifically, we consider the Leuven Natural Concept Database \cite{DeDeyne2008}, which contains feature data and familiarity ratings for animal classes (e.g., ``cat'', ``chicken'', etc.).
These data were collected from Dutch speakers, and then translated to English. We follow~\citeA{Kemp2010}, who considered 113 animal classes and 757 features from this database, and for each feature $u$ and class $c$ estimated the conditional probability $p(u|c)$ based on the number of participants who generated this feature for that class (see \figref{fig:animals}B for examples). We take $\cU$ to be the set of animal features, and assume each animal class is mentally represented by the distribution it induces over features, i.e. $m_c(u) = p(u|c)$, as estimated by~\citeA{Kemp2010}. In addition, we follow \citeA{Kemp2010} in using a familiarity-based prior over animal classes, in which the probability of a class is proportional to its familiarity score. We define the need distribution to be this prior.

Given these components, we estimated the theoretical limit for animal naming (\figref{fig:animals}C) using the same method as before, this time with $3000$ values of $\beta\in[0,2^{13}]$. We then selected the most informative systems with $k=2,3,4$ categories. The number of categories, $k$, was determined by considering categories $w$ with probability mass $\qb(w)>0.00001$. These systems are shown in \figref{fig:animals}D, where each layer of the hierarchy corresponds to a system and each box corresponds to a category within that system.
The top layer, with a single category, corresponds to a non-informative system that does not distinguish between different animal classes. This can be considered as a stage 1 system in Brown's sequence. The second layer (shown in orange) roughly corresponds to a stage 2 system. It consists of a \textit{fish} category, as can be inferred from the distribution it induces over features and animals, and another category for all other animals. It lies very close to the origin in~\figref{fig:animals}C, as it maintains little information about most animals. The third layer (shown in red) corresponds to a system with categories for \textit{fish} and \textit{wug}, as well as a category that is dominated by birds and mammals. The \textit{bird-mammal} category has greater probability mass ($0.8$) than the \textit{wug} category ($0.14$), suggesting that it is more prominent even though these two categories appear together.  This transition deviates from Brown's sequence in the early appearance of \textit{wug} (although not strongly weighted here), and in lacking a \textit{snake} category (although animals from that category do appear in the Leuven database).
One possible explanation for this deviation is that the feature data on which we relied were obtained from Dutch participants, and are thus strongly biased toward Western societies.
In the next layer (shown in blue), the 3-category system evolved to a 4-category system by refining the \textit{bird-mammal} category, resulting in a system that roughly corresponds to a Brown stage 6 system, with the exception of \textit{snake}.

These results suggest that animal naming systems may evolve under efficiency pressure much as color appears to, despite the qualitative difference between these domains. However, in order to test this proposal more comprehensively, fine-grained cross-linguistic animal naming data is required, comparable to the naming data for colors and containers.
The fact that systems along the theoretical limit capture some cross-linguistic tendencies in animal taxonomies is notable, given that our characterization of the domain, in terms of features, was necessarily strongly biased toward animal representations in Western societies.
This finding supports the idea that to some extent at least there is a shared underlying representation of animals across cultures~\cite{Mayr1969}, while also raising the interesting possibility of some cross-language and cross-cultural differences in underlying representations. 
It is also worth noting that the salient features in the IB systems tend to be both perceptual (e.g., ``is big'') and functional (e.g., ``is edible''), suggesting that both types of features may shape animal categories across languages, and that this may be consistent with pressure for efficiency~\cite{Kemp2018}.

Although we introduced the hierarchy in \figref{fig:animals}D as an account of category structure across languages, the same hierarchy could potentially serve as a model of hierarchical structure within a single language. This within-language interpretation resembles previous applications of the IB principle to language~\cite{Pereira1993}, although these applications were based on corpus statistics. The within-language interpretation seems useful in the case of animal taxonomies, a semantic domain with strong hierarchical structure, as opposed to containers and even colors.
A possible, yet speculative, reconciliation of the within-language and cross-language interpretations is that speakers may internally represent a hierarchy induced by an evolutionary sequence. For example, \citeA{Boster1986} showed that English speakers can recapitulate Berlin and Kay's implicational color hierarchy in a sequential pile-sorting task. Thus, it seems at least possible that a similar phenomenon may also hold for animal categories.

\section{General discussion}

Artifacts, animals, and colors are qualitatively different elements of human experience, yet our findings suggest that their semantic representations across languages is governed by the same general information-theoretic principle: efficient coding of meanings into words, as defined by the IB principle. We have shown that this theoretical account, which was previously tested only in the domain of color naming (ZKRT), generalizes to container names and animal taxonomies. This finding resonates with the proposal that word meanings may be shaped by pressure for efficient communication~\cite{Kemp2018}. However, it goes beyond that proposal by explaining how pressure for efficiency may account for soft categories and inconsistent naming, both in monolinguals and bilinguals, and how it may relate to language evolution.

An important direction for future research is to test to what extent our results extend to other semantic domains, and ideally, to the lexicon as a whole. While it may not be possible to apply this approach to every aspect of the lexicon, we believe that the theoretical formulation considered here may be broadly applicable across semantic domains.

\section{Acknowledgments}

We thank Anne White, Gert Storms, and Barbara Malt for making their container naming and sorting data publicly available. The animal features and familiarity data we used were preprocessed by~\citeA{Kemp2010}. We thank Simon De Deyne for initially sharing these data, and for useful discussions.
This study was partially supported by the Gatsby Charitable Foundation (N.Z. and N.T.), and by the Defense Threat Reduction Agency (N.Z. and T.R.); the content of the study does not necessarily reflect the position or policy of the U.S. government, and no official endorsement should be inferred.

\balance

\bibliographystyle{apacite}
\setlength{\bibleftmargin}{.125in}
\setlength{\bibindent}{-\bibleftmargin}
\bibliography{bibliography}

\end{document}